\documentclass[a4paper,11pt]{article}

\usepackage{fourier}
\usepackage{color}
\usepackage{graphicx}
\usepackage{url}
\usepackage[affil-it]{authblk}
\usepackage{amsmath}
\usepackage{wrapfig}

\usepackage[T1]{fontenc}
\usepackage{times}

\usepackage{tabularx} 
\usepackage{arydshln} 
\usepackage{multirow} 
\usepackage{mathtools} 
\usepackage{textcomp} 
\usepackage{makecell} 

\usepackage{enumitem} 
\usepackage{cite}

\usepackage{hyperref}
\usepackage[capitalize]{cleveref}
\crefname{section}{Sec.}{Secs.}
\Crefname{section}{Section}{Sections}
\Crefname{table}{Table}{Tables}
\crefname{table}{Tab.}{Tabs.}

\usepackage{float} 
\usepackage{caption} 

\begin{document}

\title{ DF-Net: The Digital Forensics Network for Image Forgery Detection}

\author{David Fischinger and Martin Boyer}
\affil{Austrian Institute of Technology}
\date{}
\maketitle
\thispagestyle{empty}


\begin{abstract}
The orchestrated manipulation of public opinion, particularly through manipulated images, often spread via online social networks (OSN), has become a serious threat to society. In this paper we introduce the Digital Forensics Net (DF-Net), a deep neural network for pixel-wise image forgery detection. The released model outperforms several state-of-the-art methods on four established benchmark datasets. Most notably, DF-Net's detection is robust against lossy image operations (\emph{e.g} resizing, compression) as they are automatically performed by social networks.  
\end{abstract}

\section{Introduction}
\label{sec:introduction}

"Fake News" poses an ever-growing challenge in our society as technological advancements facilitate the production of high-quality forgeries in digital media like audio, video, and images. This impact ranges from satirical memes to orchestrated political Fake News campaigns that aim to manipulate public opinion. In this paper, we introduce an effective approach to identify manipulated regions in images. This enables institutions such as media organizations and interested citizens to get a better indication of whether specific images may have been manipulated.

Over the past decade, various methods have been proposed to detect different categories of image forgery, including: copy-move, splicing, inpainting, and various enhancement techniques. However, these approaches often concentrate on specific features of each manipulation type. In recent years, more general approaches for multiple manipulation types were developed, such as \cite{Wu2019} and \cite{Wu2022}. Each of them promotes sophisticated and problem-specific network architectures and concepts, like modeling known and unknown noise on images that result from transmission to Online Social Networks (OSN).
\begin{figure}[t]
    \resizebox{0.9\columnwidth}{!}{ 
    \newcolumntype{Y}{>{\centering\arraybackslash}m{3cm}}
    \begin{tabularx}{\columnwidth} {Y Y Y Y Y}
        & \includegraphics[width=1.0\linewidth]{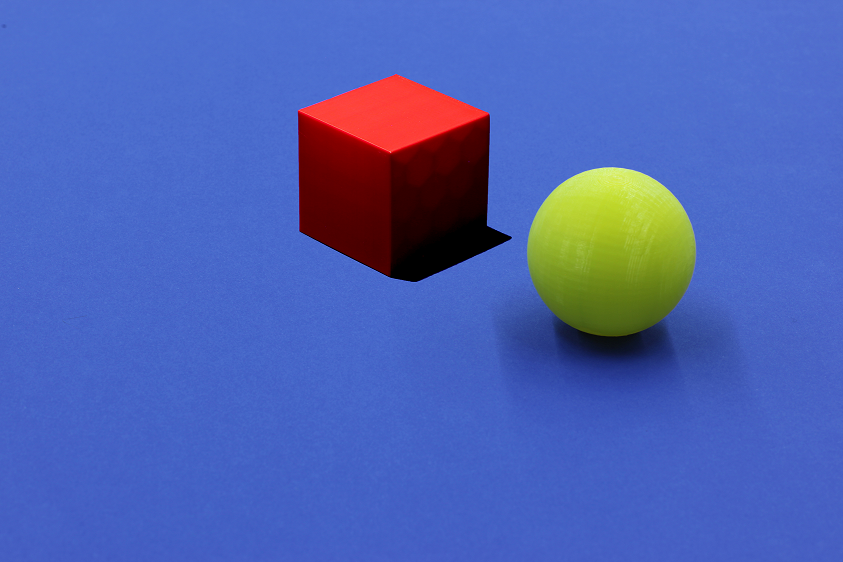} &  
        \includegraphics[width=1.0\linewidth]{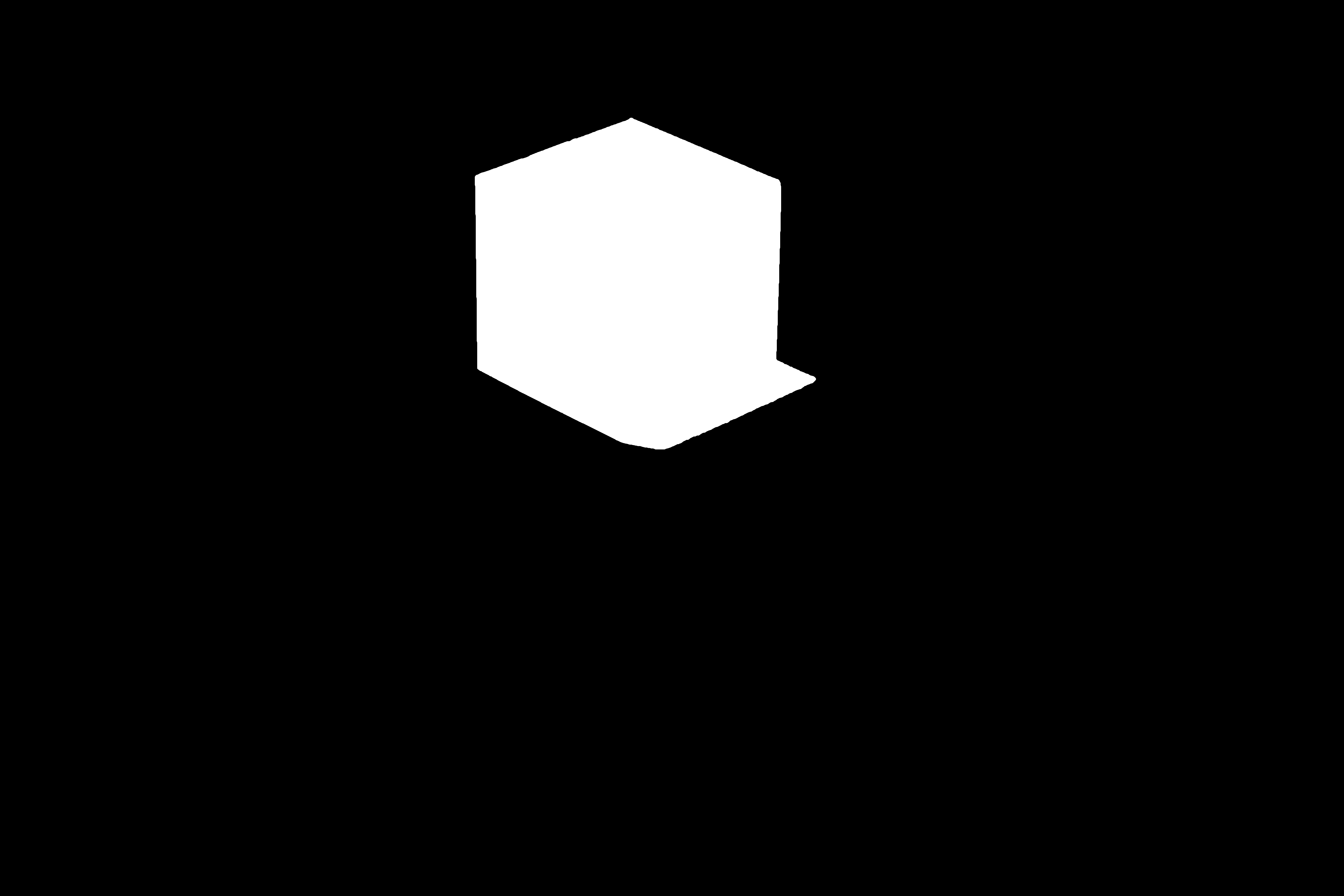} &
        \includegraphics[width=1.0\linewidth]{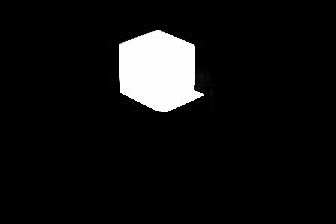} &\\
        & \includegraphics[width=1.0\linewidth]{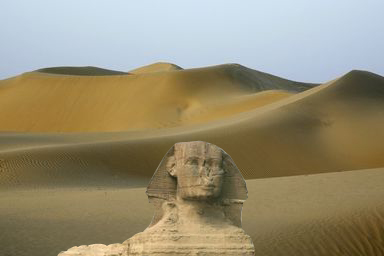} &  
        \includegraphics[width=1.0\linewidth]{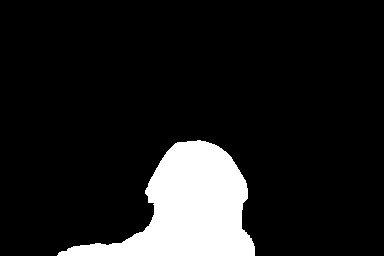} &
        \includegraphics[width=1.0\linewidth]{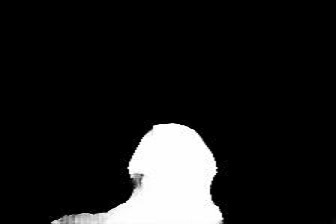} &\\
        & Forged Image & 
        Ground Truth & 
        Result & \\
    \end{tabularx}
    }
    \caption{Forgery detection results of our network. Example images are taken from the \texttt{CASIA} \cite{CASIA} and the \texttt{NIST} \cite{NIST} datasets.}
    \label{tab:examples-introduction}
\end{figure}

In this paper we present the DF-Net, an image forgery detector trained on the DF2023 dataset \cite{Fischinger2023DF2023}. To be more specific, our main contributions are as follows:\vspace*{-0.5\baselineskip}
\begin{itemize}[leftmargin=*]
     \item \textbf{Model:} Our proposed forgery detection model outperforms several state-of-the-art methods. We show its evaluation on four benchmark datasets. The model's deep learning network architecture combines the strengths of two specialized sub-models, trained from scratch on our DF2023 dataset. \vspace*{-0.65\baselineskip}
    \item \textbf{OSN robustness:} We show the robustness of our model against lossy operations (e.g. resizing, compression) as automatically done by online social networks in an extensive evaluation.\vspace*{-0.65\baselineskip}
    \item \textbf{Speed:} We present a processing time comparison with a SOTA approach that shows a significant reduction in time, especially for larger images.
\end{itemize}

\section{Related Work}
\label{sec:relatedwork}
Many methods of detecting and localizing image forgery were published (see, for example, the review of \cite{Verdoliva2020} and references therein) in order to ensure visual information authenticity. Some of these forensic techniques are designed to detect specific forms of tampering, such as splicing \cite{Lyu2013}, copy-move \cite{Wang2017, Mahmood2017, Ouyang2019, Zedan2021,Zhong2020}, and inpainting \cite{Li2017}. Unfortunately, these forensic approaches can only be applied to detect specific tampering manipulations.

In recent years, deep learning-based methods were developed to address the problem of detecting general (compound) types of forgeries. Notably, \cite{Wu2019} proposes a unified deep neural architecture called ManTra-Net, which is an end-to-end network that performs both detection and localization without extra preprocessing and postprocessing. ManTra-Net is a fully convolutional network which can handle images of arbitrary sizes and many known -- and even unkown -- forgery types. Furthermore, the authors design a self-supervised learning task to learn robust image manipulation features, formulate the forgery localization problem as a local anomaly detection problem, and propose a long short-term memory (LSTM) solution to assess local anomalies. 

The work of \cite{Zhuang2021} addresses the issue of tampering localization by focusing on the detection of commonly used editing tools and operations in Photoshop. A fully convolutional encoder-decoder architecture is designed, as well as a training data generation strategy by resorting to Photoshop scripting. 

The widespread availability of online social networks (OSN), \emph{e.g.}, Twitter, Facebook, Whatsapp, etc., makes them the dominant channels for transmitting forged images. However, almost all OSN manipulate the uploaded images in a lossy fashion (including format conversion, resizing, enhancement filtering and JPEG compression). The noise introduced by these lossy operations could severely affect the effectiveness of forensic methods. In a recent paper by \cite{Wu2022}, the problem of OSN-shared image forgeries is tackled by employing a dedicated training scheme. A baseline detector is presented, which is based on a modified U-Net \cite{Ronneberger2015} as the backbone architecture. Next, an analysis of the noise introduced by OSN is conducted, and the noise is decoupled into two parts, i.e., predictable noise and unseen noise. These are then modelled separately and the modelled noise is further incorporated into the training framework.

\textit{Outline:} The rest of this paper is structured as follows: In section \ref{sec:networkarchitecture}, we present the DF-Net, evaluate different model design choices and investigate combinations of the models. In section \ref{sec:evaluation}, our proposed model is evaluated and compared to state-of-the-art methods, specifically for OSN transmitted images. Final remarks are made in section \ref{sec:conclusion}.

\section{Network Architecture}
\label{sec:networkarchitecture}

\subsection{Architecture} \label{sec:architecture}
The DF-Net (available from \emph{https://zenodo.org/record/8142658}) is designed to detect and localize image forgeries of various types. Essentially, this is a binary segmentation problem in which each pixel of an image is classified as either pristine or forged, resulting in a binary mask~M. 

Our proposed network comprises two sub-networks (M1, M2). Both networks use U-Net \cite{Ronneberger2015} implementations, an architecture commonly used in the area of image segmentation. 

The U-Net architecture of M1 is depicted in \cref{fig:network-architecture}: U-Nets are Convolutional Neural Networks (CNNs) which consist of an encoding part where the spatial dimensions are downscaled (downsampling), and a decoding part where the spatial dimensions are increased (upsampling). On the first four sampling stages, skip connections are used which provide multi-channel feature maps from the encoding part directly to the decoding stage with the same spatial dimensions. Our U-Net implementation takes RGB images of size (256,256,3) as input. In each scaling step, we use two times a building block consisting of a 3x3 convolution, a batch normalization layer and a Relu activation layer, followed by a spatial and channel Squeeze \& Excitation (scSE) block \cite{Roy2018} and a (2x2) Max-pooling (downscaling) or a (3x3) Conv2DTranspose layer (upscaling). In the upscaling phase, the skip features are concatenated with the output of the Conv2DTranspose layer. The scSE layer can be seen as a re-calibration method for the network with a relatively small overhead regarding computing resources. During the training process, these blocks amplify spatial areas and channels which contribute more to better solutions and diminish the influence of worse performing network parts. At the final layer of the network, a 1x1 convolution with sigmoid activation function is used to calculate a value in $[0,1]$ which indicates how likely each pixel is manipulated.

    \begin{table}[!htb]
    \begin{minipage}{.545\linewidth}
        \centering
          \includegraphics[width=1.0\columnwidth]{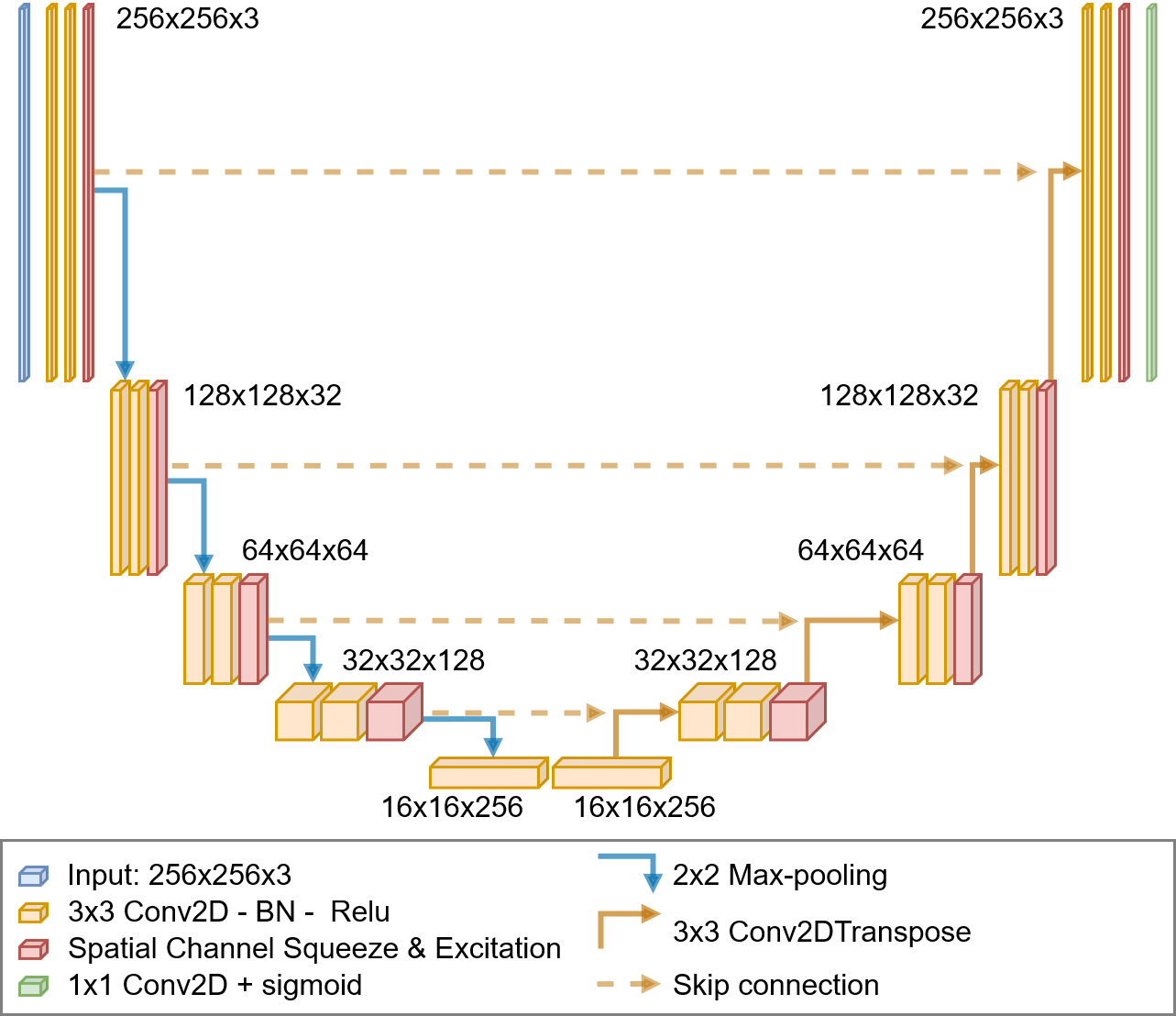}
          \captionof{figure}{Network architecture of submodel M1: A \mbox{U-Net} architecture with 4 skip connections and spatial channel Squeeze \& Excitation (scSE) extension. A more detailed description can be found in section \ref{sec:architecture}. }
          \label{fig:network-architecture}
    \end{minipage}
    \begin{minipage}{.03\linewidth}
    \ \ 
    \end{minipage}
    \begin{minipage}{.425\linewidth}
      \centering
          \includegraphics[width=1.0\columnwidth]{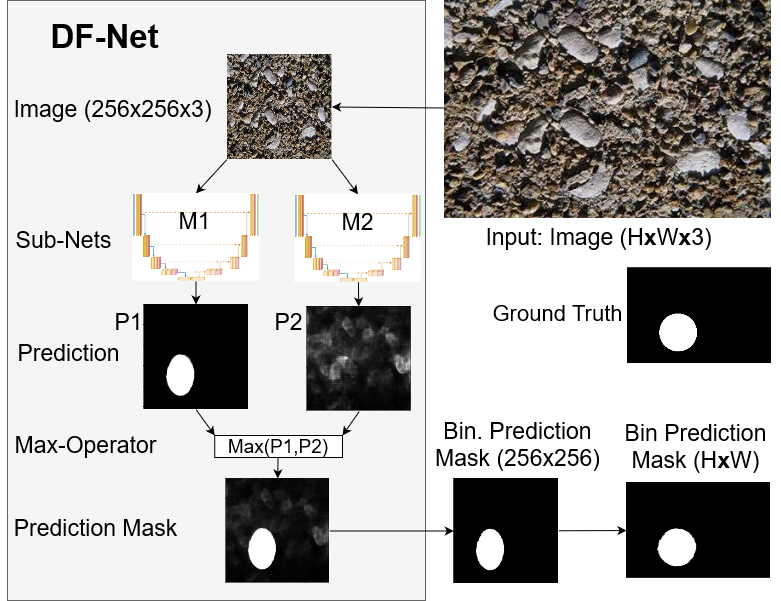}
          \captionof{figure}{Network architecture of \mbox{DF-Net}: Example of Model combination for image Sp\_S\_NNN\_C\_txt0019\_txt0019\_0019.jpg from the CASIA\_V1 dataset.}
          \label{fig:dfnetwork-architecture}
    \end{minipage}
    \end{table}


The architecture of model M2 slightly deviates from M1: for each Conv2D block of M2, the kernel size is set to 5x5 instead of 3x3 for exactly 4 filters. The output feature maps of the filters with different-sized kernels are concatenated again before the batch normalization is done.

We listed major evolutionary steps of the DF-Net in \cref{tab:subnetcomparison}, where we compared networks on four benchmark datasets. Considerable performance boosts were accomplished when adding the spatial channel Squeeze \& Excitation (scSE) calibration blocks (plus of $0.082$) and after the input size for training and prediction was increased from $(224x224)$ to $(256x256)$ (0.029). Model M2 was only trained on splicing images from DF2023, but still performed best out of all single sub-networks. The architectural modification replacing four $3x3$ kernels with $5x5$ kernels in each convolution gave an additional boost of $0.014$. A considerable performance gain was reached by combining networks. With the maximum operator, better results were achieved compared to averaging the predictions of two single networks. Taking the maximum prediction value from M1 and M2 for each image pixel, resulted in an average value of $0.583$ which exceeds the best performing sub-model $M2$, which only achieved $0.535$ overall. A deeper analysis showed that the maximum-operator-based model performs worse than the better sub-model, for most images. For the CASIA dataset, the AUC value for max(M1,M2) is higher than the better performing submodule for only 86 out of 920 images. But the AUC performance is considerably higher than the average AUC value of the two sub-modules. In other words: By applying the maximum operator, the models prediction on each image is generally almost as good as the better-performing sub-module. So, the final DF-Net architecture depicted in \cref{fig:dfnetwork-architecture} combines the strengths of the separately trained models M1 and M2 by taking the maximum prediction value per pixel from the outputs of both sub-nets.

\begin{table*}[!ht]
    \centering
    \resizebox{\textwidth}{!}{
        \begin{tabular}{|l|l|l|l|l||l|l|l||l|l|l||l|l|l||l|l|l|}
            \hline \hline
             ~ & \multicolumn{4}{c}{} & \multicolumn{12}{c|}{Test Datasets}\\ \cline{6-17}
             Model Arch. & 
                \multicolumn{4}{c}{Average(metrics)} &
                \multicolumn{3}{c}{\texttt{CASIA} } &
                \multicolumn{3}{c}{\texttt{Columbia} } & 
                \multicolumn{3}{c}{\texttt{DSO} } &
                \multicolumn{3}{c|}{\texttt{NIST} } \\  \cline{2-17}
                ~ & Mean & AUC & F1 & IoU & AUC & F1 & IoU & AUC & F1 & IoU & AUC & F1 & IoU & AUC & F1 & IoU \\ \hline
            U-Net (224²) & \textbf{.390} & .701 & .263 & .206 & .75 & .21 & .18 & .80 & .52 & .41 & .60 & .10 & .06 & .65 & .22 & .17 \\ \hline
            M1 (224²) & \textbf{.472} & .797 & .337 & .282 & .85 & .43 & .37 & .83 & .50 & .43 & .73 & .15 & .11 & .78 & .26 & .22 \\ \hline
            M1 (256²) & \textbf{.501} & .782 & .396 & .326 & .85 & .53 & .47 & .79 & .54 & .45 & .71 & .24 & .16 & .77 & .28 & .23 \\ \hline
            M2: M1-Ar. & \textbf{.521} & .790 & .425 & .348 & .82 & .42 & .36 & .87 & .70 & .61 & .74 & .32 & .22 & .74 & .26 & .20 \\ \hline
            M2 & \textbf{.535} & .804 & .439 & .363 & .82 & .43 & .37 & .89 & .74 & .67 & .76 & .31 & .21 & .75 & .27 & .20 \\ \hline
            avg(M1,M2) & \textbf{.57}4 & .842 & .474 & .405 & .91 & .60 & .54 & .88 & .69 & .63 & .78 & .30 & .21 & .80 & .30 & .24 \\ \hline
            max(M1,M2) & \textbf{.583} & .837 & .501 & .411 & .91 & .59 & .50 & .88 & .76 & .68 & .77 & .36 & .25 & .79 & .30 & .23 \\ \hline \hline
        \end{tabular}
    }
    \caption{Comparison of different network implementations we have trained and evaluated. The first two networks were trained on images of size $224x224$, all other networks on images of size $256x256$. Column \textbf{Mean} shows the average of the AUC, F1 and IoU metrics over all 4 datasets. The last row shows results from the DF-Net. More details are given in \cref{sec:architecture}}
    \label{tab:subnetcomparison}
\end{table*}


Our experiments show that the combination of separately trained sub-networks results in a considerable performance improvement (see \cref{tab:subnetcomparison}). Equally beneficiary is the advantage of splitting the whole model to sub-models that can be trained separately. This reduces time because researchers can faster assess if changes in the network architecture, the training data or the training parameters should be discarded or pursued. Furthermore, it allows to overcome hardware limitations which we see as a major advantage of this architecture.

\subsection{Implementation Details}
Model specification and training were done in the deep learning framework Tensorflow. 
For training and detection, the images were resized to $(256,256)$ pixels. An \texttt{Nvidia GeForce GTX 1080 Ti} GPU with 11G memory was used for training, with batch size set to $32$. We used the Adam optimizer \cite{kingma2014method} and performed $1000$ steps per epoch and stopped after the loss value of the validation dataset did not improve for $35$ epochs. Training starts with a learning rate of $0.0001$, which is halved after $10$ epochs without improvement. M1 was first trained on all manipulation types and images. Subsequently it was refined by training on the $200.000$ copy-move forgeries from DF2023 \cite{Fischinger2023DF2023}. Model M2 was trained only on the 400K splicing images from DF2023.

\section{Evaluation}
\label{sec:evaluation}

We compare the DF-Net to four state-of-the-art methods: ForSim \cite{Mayer2019ForSim}, DFCN \cite{Zhuang2021}, ManTra-Net \cite{Wu2019} and the work of Wu \textit{et al.} \cite{Wu2022}, abbreviated below as Wu22. The approaches are evaluated on the four benchmark datasets \texttt{CASIA\_V1} \cite{CASIA}, \texttt{Columbia} \cite{Columbia}, \texttt{DSO} \cite{DSO} and \texttt{NIST16} \cite{NIST}. See \Cref{tab:time-comparison} for an overview of the datsets.

\begin{table}[!h]
    \centering
    \resizebox{.95\columnwidth}{!}{
    \begin{tabular}{l c c r r}
    \hline \hline
        \textbf{Dataset} & \textbf{\#\ Images} & \textbf{Format} & \textbf{t-WU22} & \textbf{ t-DF-N} \\ \hline
        \texttt{CASIA} \cite{CASIA} & 920 & jpg & 169 & \textbf{155} \\
        \texttt{Columbia} \cite{Columbia} & 160 & tif & 120 & \textbf{28} \\ 
        \texttt{DSO} \cite{DSO} & 100 & png & 701 & \textbf{27} \\ 
        \texttt{NIST} \cite{NIST} & 564 & jpg & 15250 & \textbf{235} \\ \hline \hline
    \end{tabular}
    }
    \caption{Benchmark datasets with processing times (t) for Wu22 \cite{Wu2022} and DF-Net (ours) in seconds for predictions per benchmark dataset. For datasets with huge images such as \texttt{NIST} (images of size up-to 5616×3744 pixels), tile-based approaches take considerably longer than approaches performing pre-scaling.}
    \label{tab:time-comparison}
\end{table}

\subsection{Online Social Networks}
The popularity of online social networks (OSN) makes them the dominating channels for the distribution of manipulated images in the context of entertainment, but also for fake news, disinformation and propaganda. Unfortunately, OSN automatically apply operations like compression and resizing, which reduce valuable information for image forgery detection. To show the robustness of DF-Net against these lossy operations, all the SOTA methods are tested against OSN adapted versions of the four benchmark datasets. In \cite{Wu2022}, the authors transmitted the images of the benchmark datasets via the social online platforms Facebook, Whatsapp, Weibo, and Wechat and made the collected datasets and their evaluation of several state-of-the-art methods available to the research community.
We could reproduce the results for Wu22 \cite{Wu2022}. For ManTra-Net \cite{Wu2019}, we tested the officially released TensorFlow model which is different from the model used for the authors' evaluation as the authors stated in \cite{ManTraNetRepo2022}, and a public PyTorch re-implementation \cite{ManTraNetPyTorch2022} as well. Here we could reproduce the ManTraNet results as stated in \cite{Wu2022}. For ForSim \cite{Mayer2019ForSim} and DFCN \cite{Zhuang2021} results from the evaluation in \cite{Wu2022} are stated in \cref{tab:osnevaluation}, together with the evaluation of the proposed DF-Net.

\subsubsection{Evaluation Criteria:} We adopt three metrics commonly used in the area of image forgery detection: Area under the receiver operating characteristic curve (AUC), F1-score and Intersection over Union (IoU). The metrics are calculated on a pixel-level. For IoU and F1-score, the threshold for the output of the trained networks is set to $0.5$.

\begin{table*}[!]
    \centering
    \resizebox{\textwidth}{!}{ 
        \begin{tabular}{|l|l|l|l|l||l|l|l||l|l|l||l|l|l||l|l|l||l|}
        \hline \hline
                ~ & ~ & \multicolumn{16}{c|}{Test Datasets}\\ \cline{3-18}
            Models & OSN &
            \multicolumn{3}{c}{\texttt{CASIA} } &
            \multicolumn{3}{c}{\texttt{Columbia} } &
            \multicolumn{3}{c}{\texttt{DSO} } &
            \multicolumn{3}{c}{\texttt{NIST} } &
            \multicolumn{4}{c|}{Average}\\ \cline{3-18}
            ~ & ~ & AUC & F1 & IoU & AUC & F1 & IoU & AUC & F1 & IoU & AUC & F1 & IoU & AUC & F1 & IoU & Mean \\ \hline
            DFCN & - & .654 & .192 & .119 & .789 & .541 & .395 & .724 & .303 & .227 & .778 & .250 & .204 & .736 & .322 & .236 & .431 \\
            FSim & - & .554 & .169 & .102 & .731 & .604 & .474 & .796 & \textbf{.487} & \textbf{.371} & .642 & .188 & .123 & .681 & .362 & .268 & .437 \\ 
            MNet & - & .776 & .130 & .086 & .747 & .357 & .258 & .795 & .344 & .253 & .634 & .088 & .054 & .738 & .230 & .163 & .377 \\ 
            Wu22 & - & .873 & .509 & .465 & .862 & .707 & .608 & \textbf{.854} & .436 & .308 & .783 & \textbf{.332} & \textbf{.255} & \textbf{.843} & .496 & .409 & .5827 \\ 
            \textbf{DF-Net} & - & \textbf{.906} & \textbf{.589} & \textbf{.496} & \textbf{.880} & \textbf{.757} & \textbf{.679} & .769 & .360 & .246 & \textbf{.793} & .299 & .226 & .837 & \textbf{.501} & \textbf{.411} & \textbf{.5832} \\ \hline
            
            DFCN & Facebook & .654 & .190 & .116 & .687 & .479 & .338 & .673 & .238 & .184 & .705 & .207 & .138 & .680 & .278 & .194 & .384 \\ 
            FSim & Facebook & .537 & .157 & .094 & .607 & .450 & .304 & .689 & .356 & .238 & .580 & .140 & .085 & .603 & .276 & .180 & .353 \\ 
            MNet & Facebook & .763 & .102 & .065 & .626 & .103 & .056 & .638 & .109 & .071 & .652 & .095 & .057 & .670 & .102 & .062 & .278 \\ 
            Wu22 & Facebook & .862 & .462 & .417 & \textbf{.883} & .714 & .611 & \textbf{.859} & \textbf{.447} & \textbf{.320} & .783 & \textbf{.329} & \textbf{.253} & \textbf{.847} & .488 & .400 & .578 \\
            \textbf{DF-Net} & Facebook & \textbf{.905} & \textbf{.587} & \textbf{.492} & .883 & \textbf{.760} & \textbf{.681} & .770 & .359 & .245 & \textbf{.795} & .304 & .229 & .838 & \textbf{.502} & \textbf{.412} & \textbf{.584} \\ \hline
            
            DFCN & Wechat & .651 & .193 & .119 & .676 & .487 & .344 & .653 & .221 & .137 & .701 & .176 & .114 & .670 & .269 & .179 & .373 \\ 
            FSim & Wechat & .532 & .153 & .091 & .650 & .496 & .354 & .564 & .247 & .147 & .581 & .136 & .082 & .582 & .258 & .168 & .336 \\ 
            MNet & Wechat & .724 & .080 & .048 & .613 & .199 & .125 & .582 & .076 & .045 & .654 & .095 & .057 & .643 & .113 & .069 & .275 \\ 
            Wu22 & Wechat & .833 & .405 & .358 & \textbf{.883} & .727 & .631 & \textbf{.823} & \textbf{.366} & \textbf{.252} & .764 & .286 & .214 & .826 & .446 & .364 & .545 \\ 
            \textbf{DF-Net} & Wechat & \textbf{.902} & \textbf{.564} & \textbf{.467} & .881 & \textbf{.759} & \textbf{.681} & .765 & .358 & .245 & \textbf{.799} & \textbf{.314} & \textbf{.238} & \textbf{.837} & \textbf{.499} & \textbf{.408} & \textbf{.581} \\ \hline
            
            DFCN & Whatsapp & .655 & .191 & .117 & .692 & .471 & .331 & .645 & .264 & .162 & .689 & .187 & .125 & .670 & .278 & .184 & .377 \\ 
            FSim & Whatsapp & .525 & .151 & .091 & .595 & .436 & .294 & .542 & .233 & .139 & .586 & .137 & .082 & .562 & .239 & .152 & .318 \\ 
            MNet & Whatsapp & .763 & .099 & .063 & .630 & .098 & .052 & .616 & .081 & .052 & .702 & .101 & .062 & .678 & .095 & .057 & .277 \\ 
            Wu22 & Whatsapp & .866 & .478 & .431 & \textbf{.889} & .727 & .628 & \textbf{.839} & .341 & .233 & .785 & .313 & .239 & \textbf{.845} & .465 & .383 & .564 \\ 
            \textbf{DF-Net} & Whatsapp & \textbf{.905} & \textbf{.588} & \textbf{.495} & .883 & \textbf{.762} & \textbf{.685} & .765 & \textbf{.361} & \textbf{.249} & \textbf{.803} & \textbf{.324} & \textbf{.247} & .839 & \textbf{.509} & \textbf{.419} & \textbf{.589} \\ \hline
            
            DFCN & Weibo & .653 & .191 & .117 & .676 & .458 & .319 & .639 & .227 & .140 & .706 & .192 & .125 & .668 & .267 & .175 & .370 \\ 
            FSim & Weibo & .542 & .165 & .100 & .610 & .453 & .312 & .568 & .260 & .165 & .581 & .150 & .094 & .575 & .257 & .168 & .333 \\ 
            MNet & Weibo & .754 & .099 & .063 & .620 & .103 & .056 & .606 & .057 & .036 & .671 & .088 & .053 & .663 & .087 & .052 & .267 \\ 
            Wu22 & Weibo & .858 & .466 & .421 & .883 & .724 & .626 & \textbf{.808} & \textbf{.370} & \textbf{.253} & .780 & .294 & .219 & .832 & .463 & .380 & .558 \\ 
            \textbf{DF-Net} & Weibo & \textbf{.902} & \textbf{.584} & \textbf{.490} & \textbf{.890} & \textbf{.766} & \textbf{.684} & .759 & .354 & .245 & \textbf{.791} & \textbf{.303} & \textbf{.230} & \textbf{.836} & \textbf{.502} & \textbf{.412} & \textbf{.583} \\ \hline \hline
        \end{tabular}
    }
    \caption{Comparison of the state-of-the-art approaches DFCN \cite{Zhuang2021}, ForSim \cite{Mayer2019ForSim} (FSim), ManTra-Net \cite{Wu2019} (MNet), Wu22 \cite{Wu2022} and our proposed DF-Net. Highest metric values per benchmark dataset and OSN are marked \textbf{bold}.}
    \label{tab:osnevaluation}
\end{table*}

\begin{figure}[!]
  \centering
  \includegraphics[width=0.5\columnwidth]{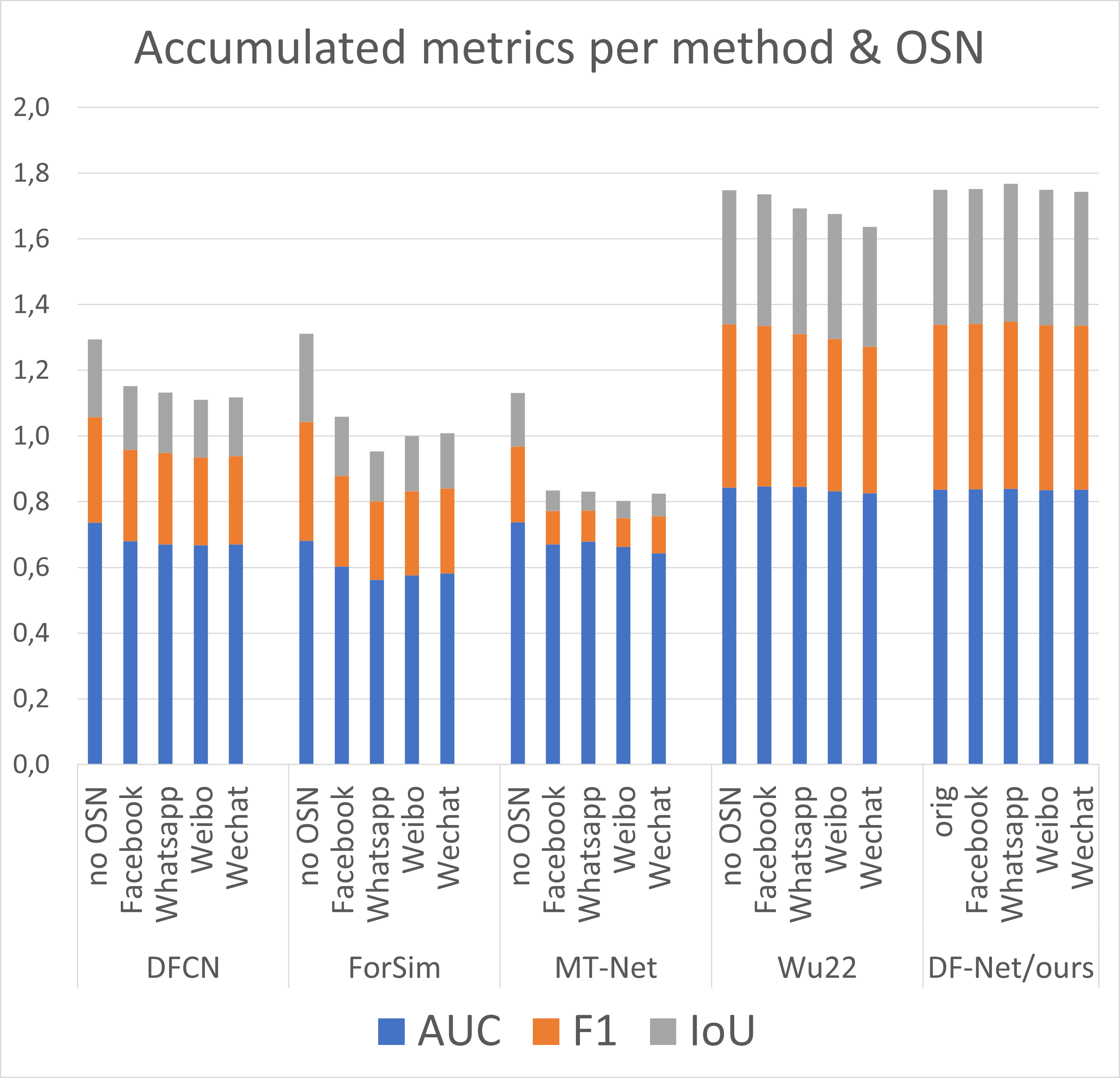}
  \caption{Metrics (AUC, F1, IoU) averaged over 4 benchmark datasets in accumulated presentation. Each column represents the combination of a method and the OSN used for dataset modification}
  \label{fig:osnresultfigure}
\end{figure}

\begin{figure*}[!]
    \centering
    \resizebox{1.0\textwidth}{!}{ 
    \newcolumntype{Y}{>{\centering\arraybackslash}X}
    \newcolumntype{Z}{>{\centering} m{1em}}
    \begin{tabularx}{\linewidth}{Z Y Y Y Y Y}
        \multicolumn{1}{c}{\rotatebox{90}{\hspace{2em}\makecell{\texttt{DSO}}}} &
        \includegraphics[width=0.99\linewidth]{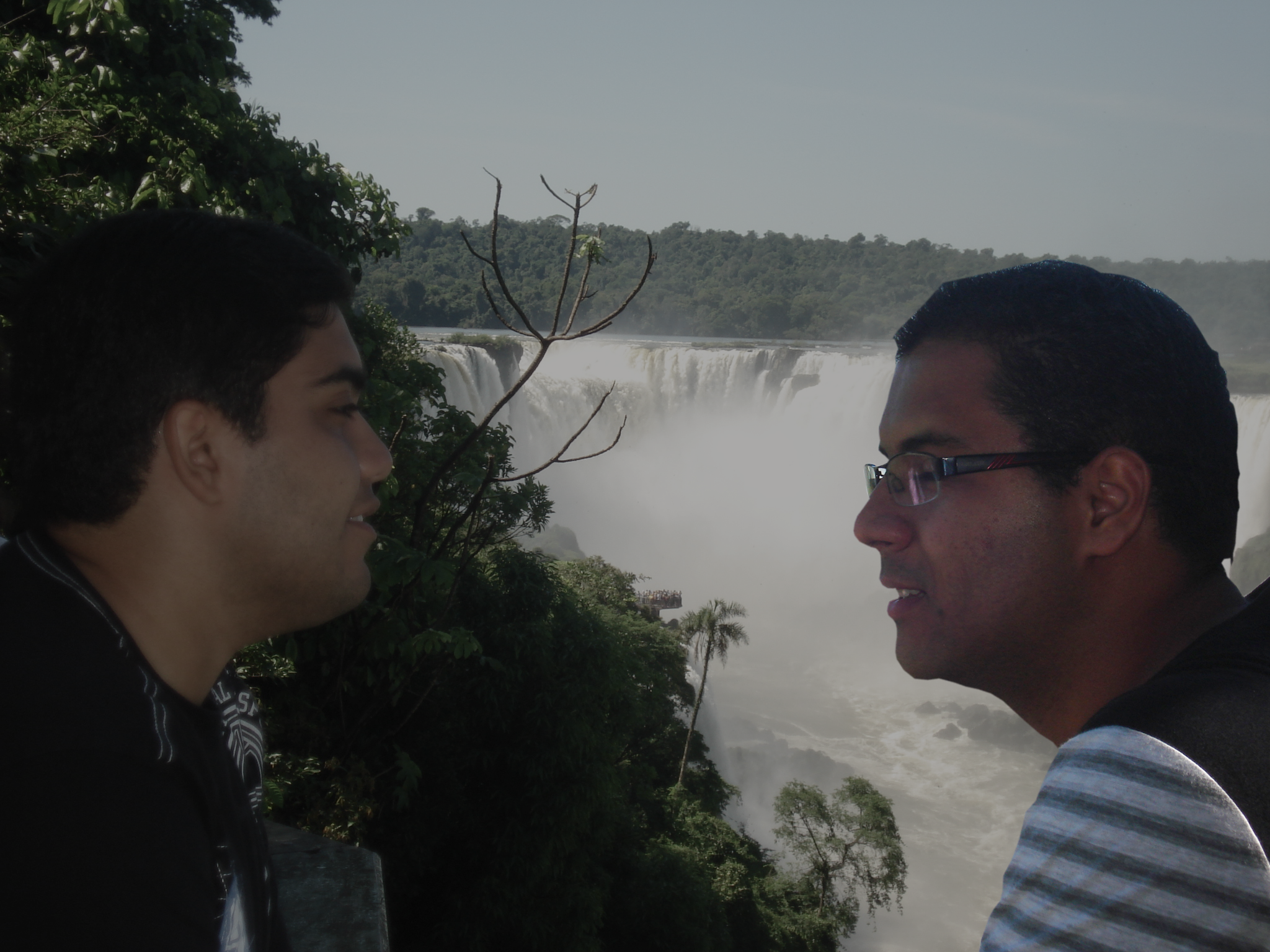} &  \includegraphics[width=0.99\linewidth]{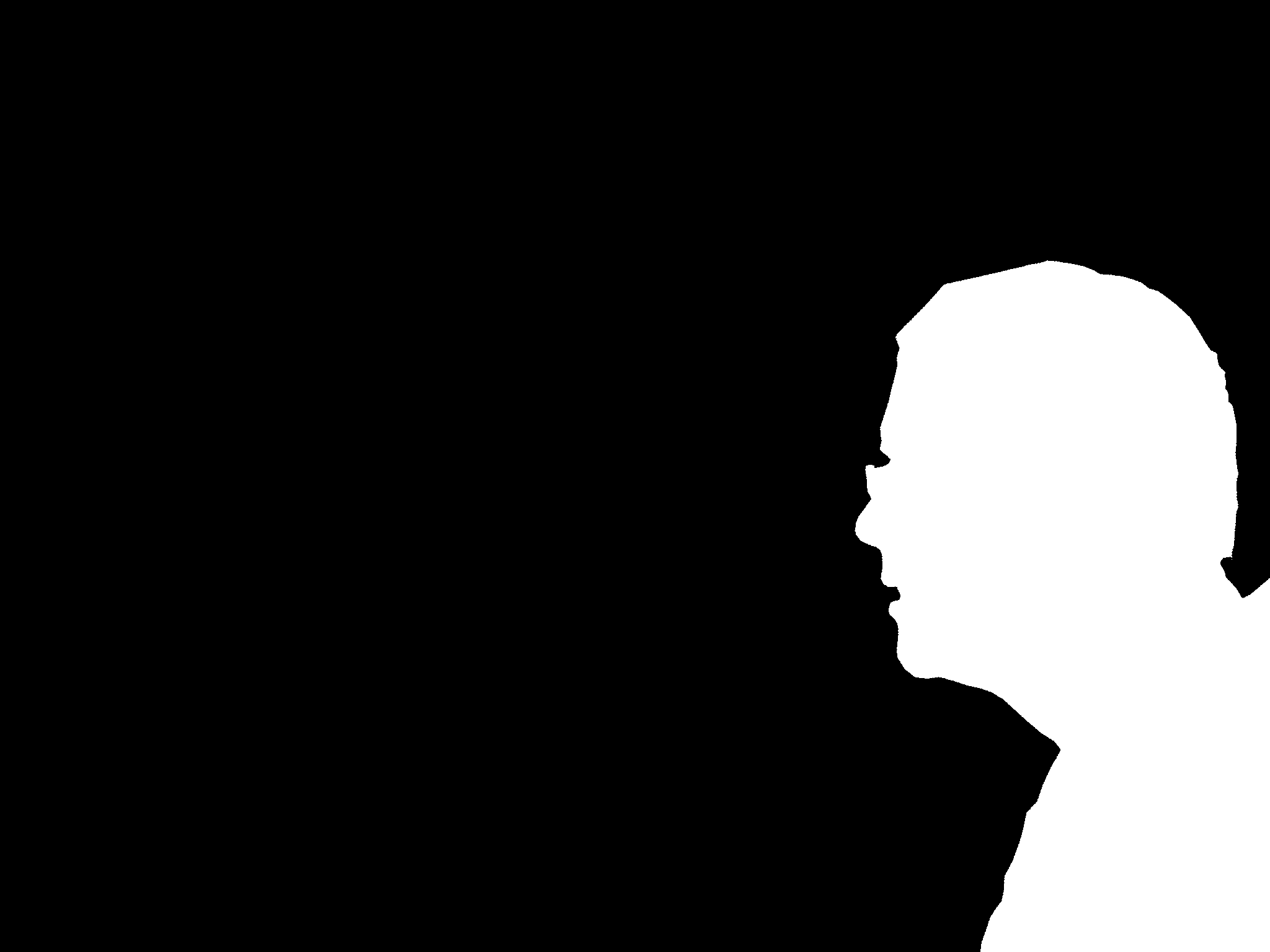} &
        \includegraphics[width=0.99\linewidth]{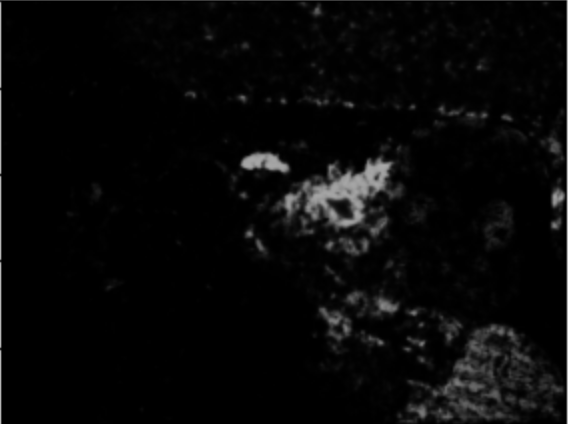} &
        \includegraphics[width=0.99\linewidth]{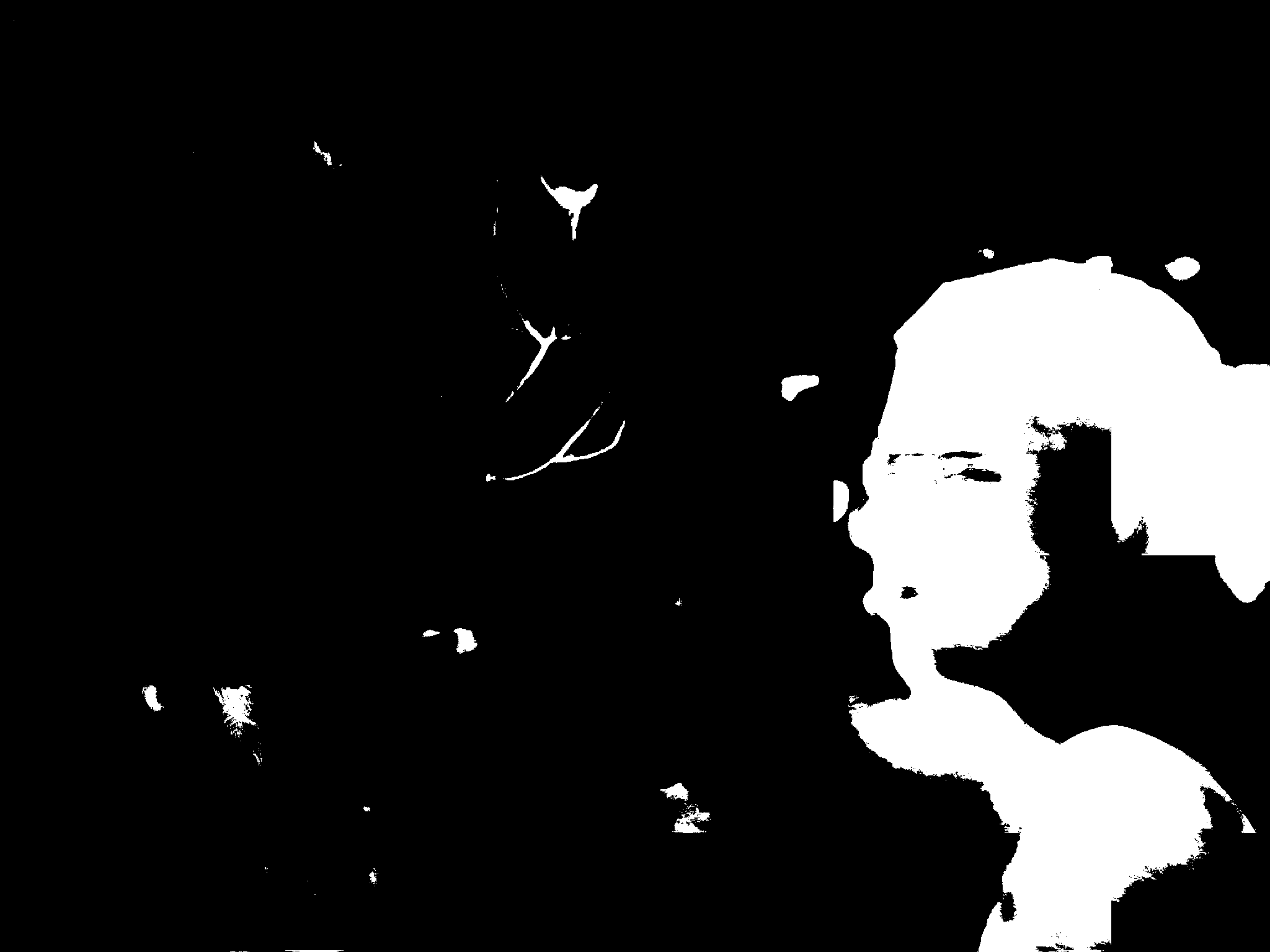} & \includegraphics[width=0.99\linewidth]{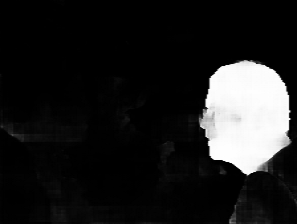} \\
        \multicolumn{1}{c}{\rotatebox{90}{\makecell{\texttt{Columbia}}}} &
        \includegraphics[width=0.99\linewidth]{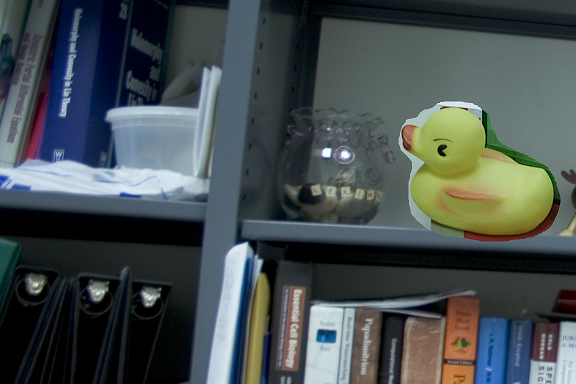} &  \includegraphics[width=0.99\linewidth]{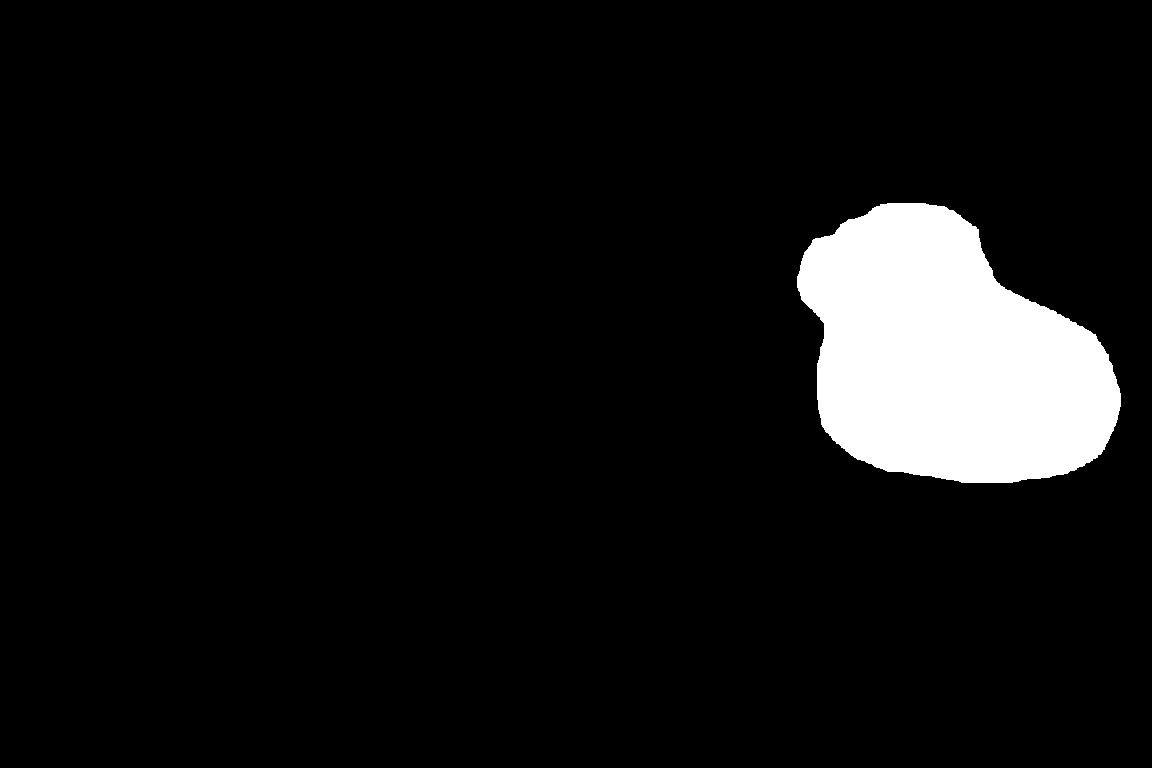} &
        \includegraphics[width=0.99\linewidth]{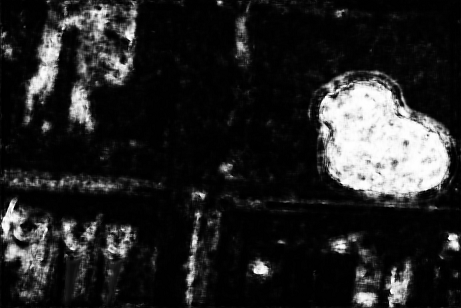} &
        \includegraphics[width=0.99\linewidth]{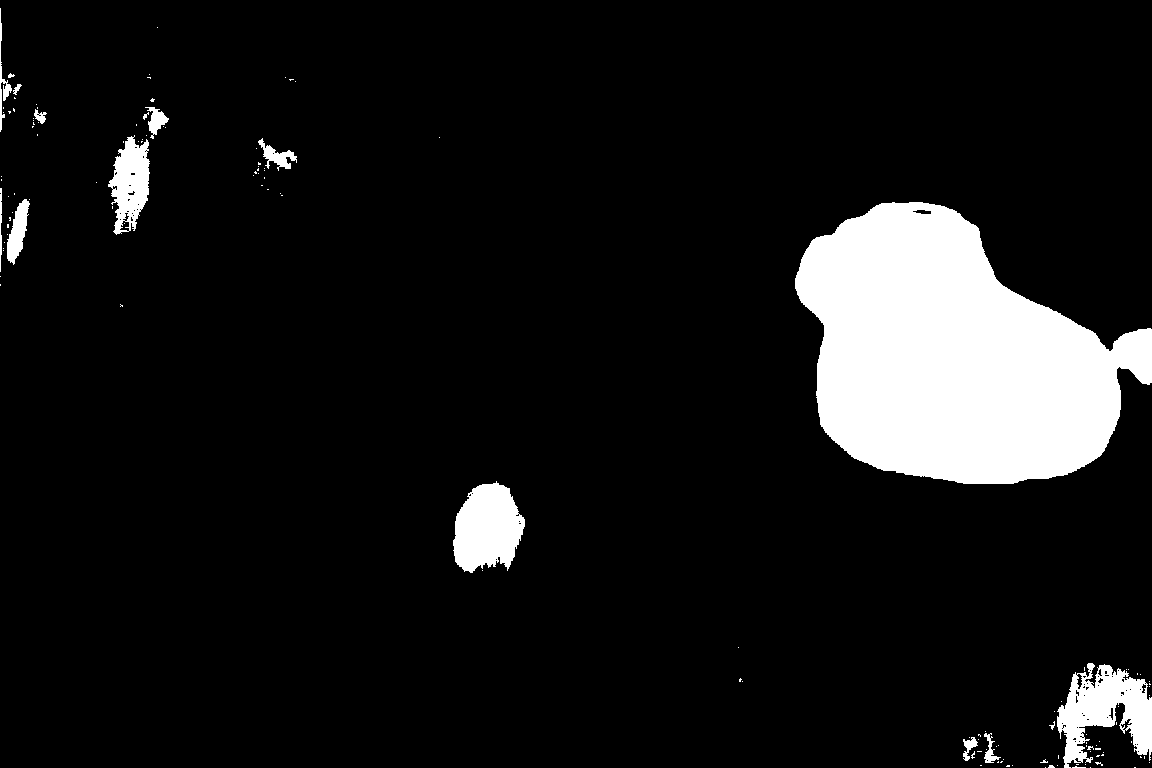} & 
        \includegraphics[width=0.99\linewidth]{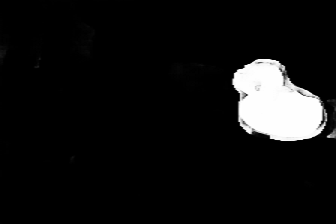} \\
        \multicolumn{1}{c}{\rotatebox{90}{\makecell{\hspace{1em}\texttt{NIST} \hspace{1em}}}} &
        \includegraphics[width=0.99\linewidth]{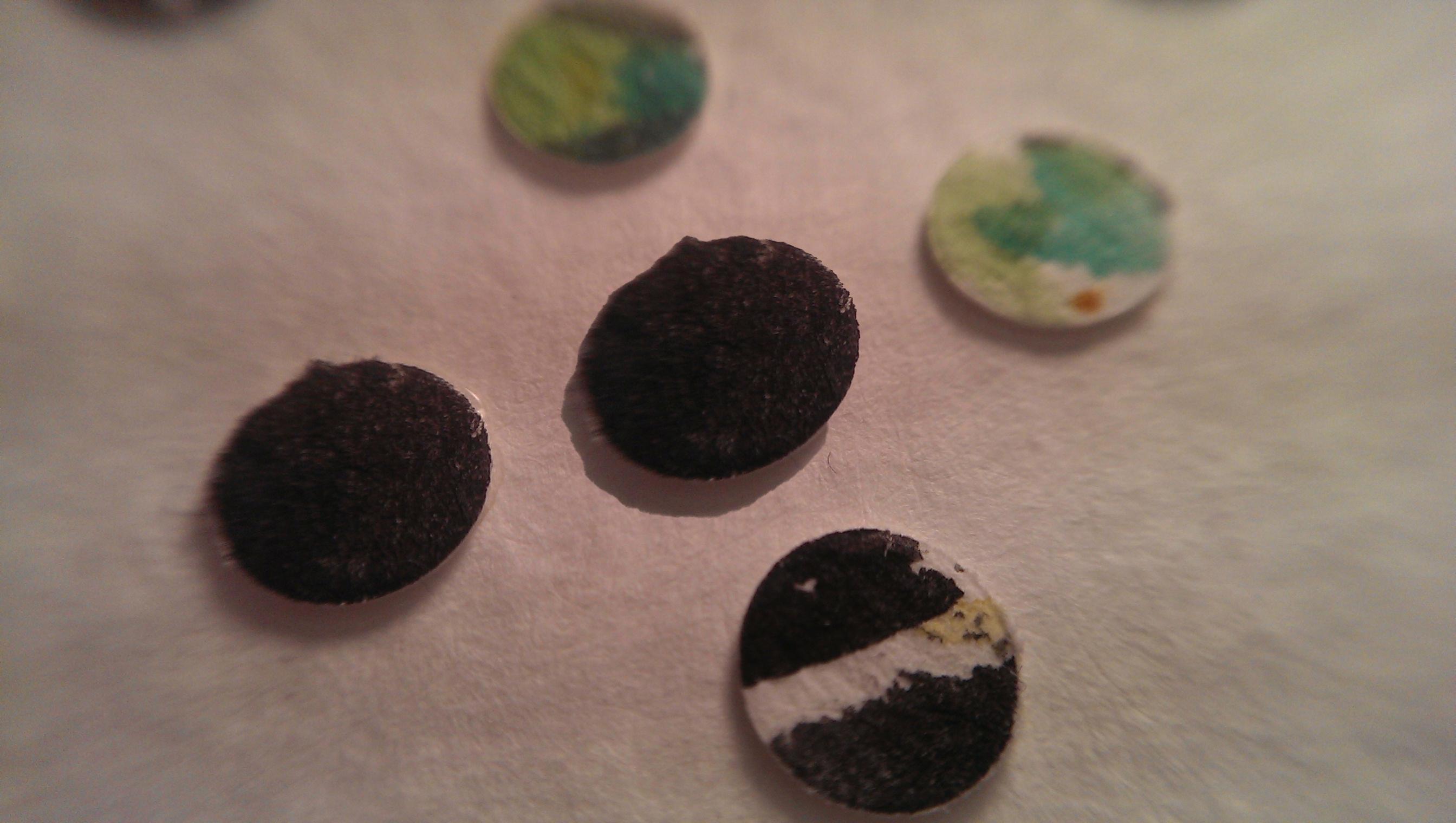} &  
        \includegraphics[width=0.99\linewidth]{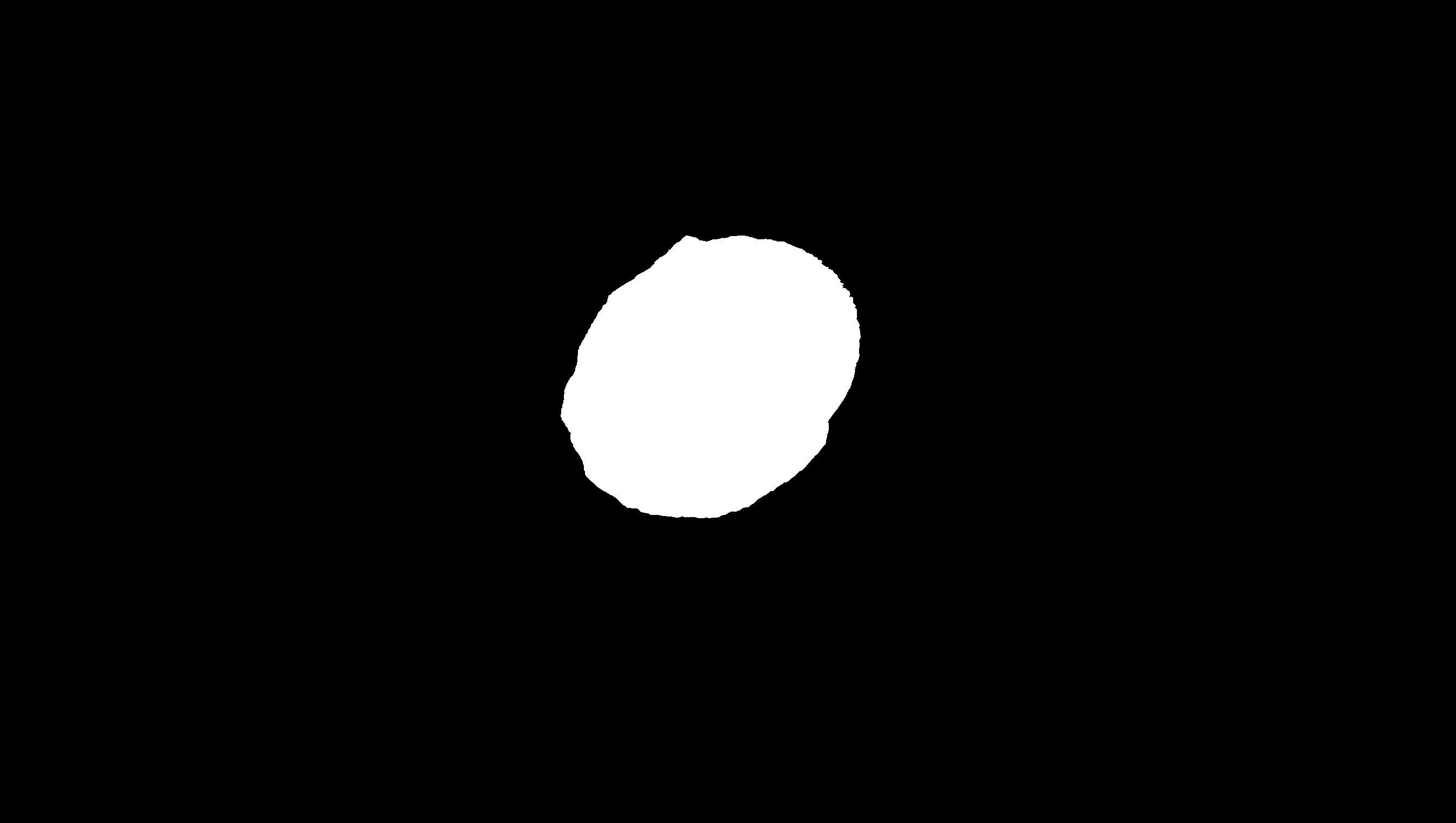} &
        \includegraphics[width=0.99\linewidth]{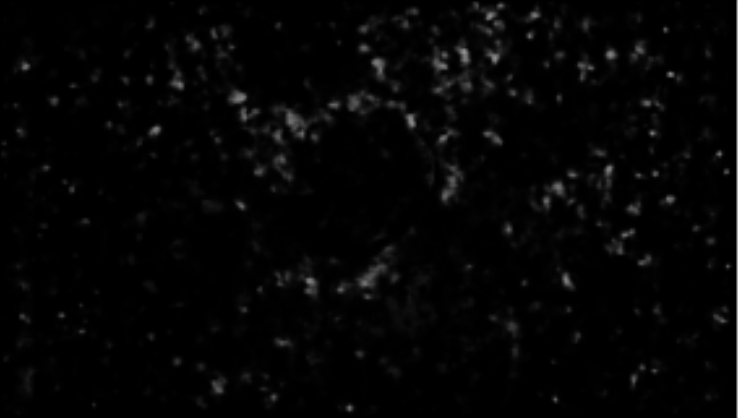} &
        \includegraphics[width=0.99\linewidth]{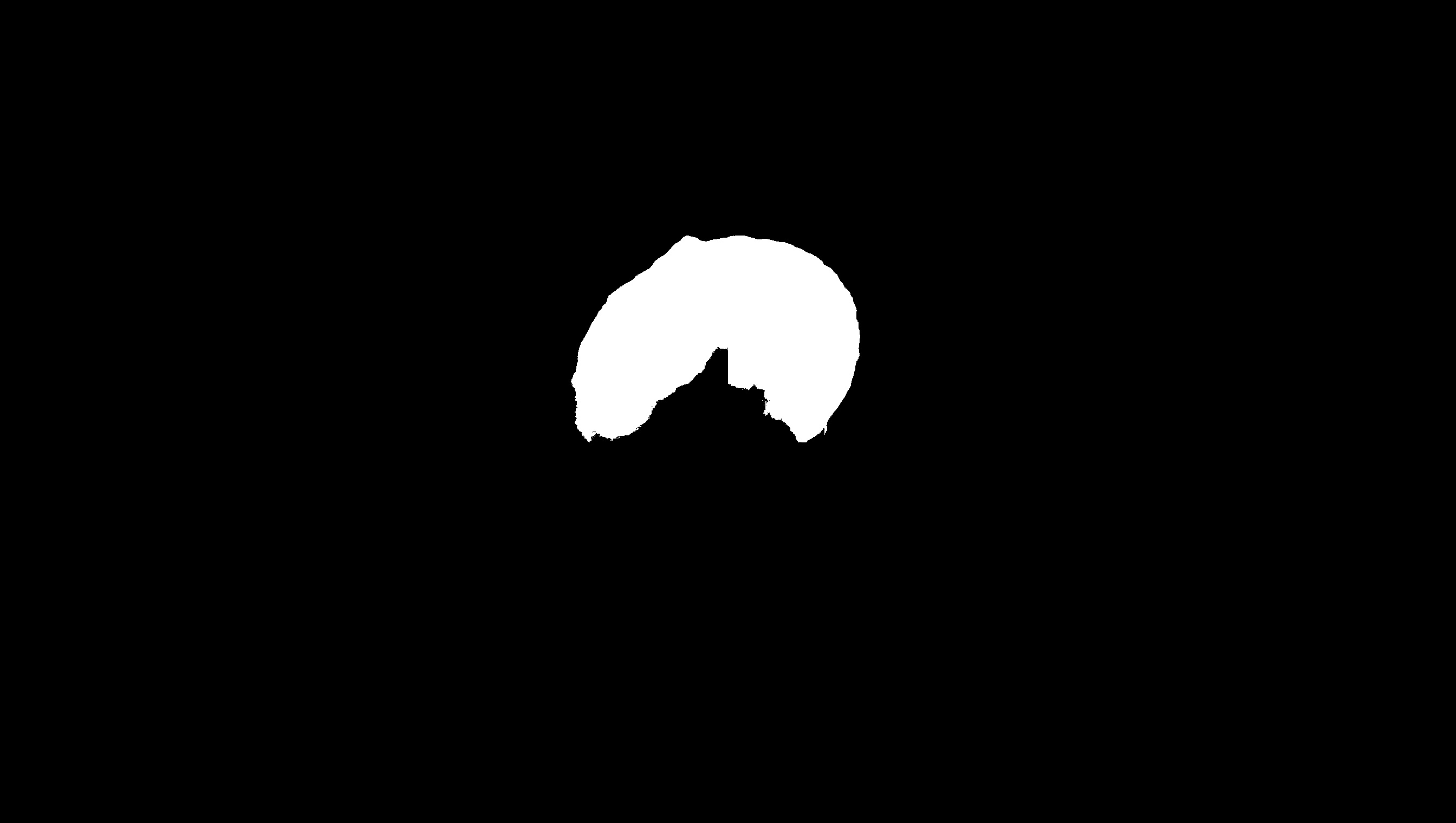} & 
        \includegraphics[width=0.99\linewidth]{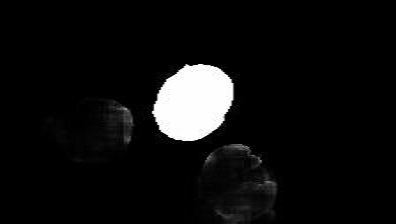} \\
        \multicolumn{1}{c}{\rotatebox{90}{\makecell{\hspace{1em}\texttt{CASIA} \hspace{1em}}}} &
        \includegraphics[width=0.99\linewidth]{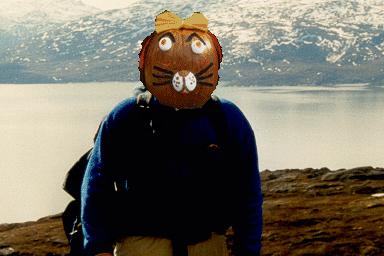} &  
        \includegraphics[width=0.99\linewidth]{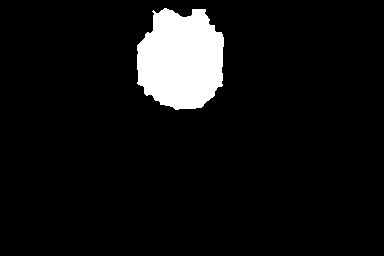} &
        \includegraphics[width=0.99\linewidth]{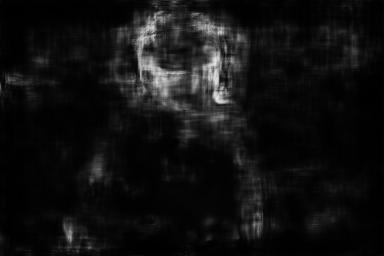} &
        \includegraphics[width=0.99\linewidth]{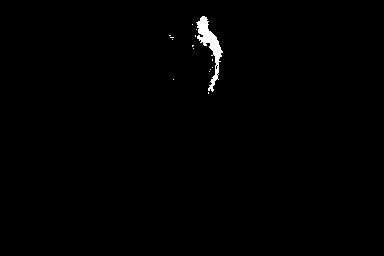} & 
        \includegraphics[width=0.99\linewidth]{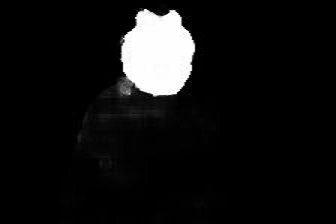} \\
        & Forged Image & Ground Truth & MantraNet & Wu22 & \textbf{DF-Net}
    \end{tabularx}
    }
    \caption{Examples of qualitative comparison of MantraNet \cite{Wu2019}, Wu22 \cite{Wu2022} and our proposed DF-Net. Each line shows one example image for each of the four benchmark datasets \texttt{DSO} \cite{DSO}, \texttt{Columbia} \cite{Columbia}, \texttt{NIST} \cite{NIST}, \texttt{CASIA} \cite{CASIA}. The five columns show: the forged image (input), manipulated area (ground truth), results (output) from MantraNet, Wu22 and DF-Net. We show example results of the M1 sub-model for the \texttt{DSO} and the \texttt{NIST} dataset.}
    \label{tab:sota-comparison}
\end{figure*}

\subsection{Quantitative Comparison}
As shown in \cref{tab:osnevaluation}, the DF-Net could clearly outperform ForSim, DFCN and ManTra-Net on the original benchmark datasets \texttt{CASIA V1}, \texttt{Columbia} and \texttt{NIST16}, and on the overall average of the metrics. The method Wu22 performed similar to the DF-Net. The overall average of DF-Net ($0.5832$) is just $0.0005$ higher. Yet the situation for the OSN modified datasets has to be noted: in \cref{fig:osnresultfigure} we visualize the sum of AUC, F1 and IoU per dataset. The methods ForSim, DFCN and ManTra-Net show a large performance decrease for the modified datasets. Robustness against OSN transmitted data was the key contribution of Wu22 \cite{Wu2022}, hence their method performs only moderately worse on OSN transmitted images compared to the original ones. DF-Net on the other hand does not show a significant performance drop at all. This can be explained by the training process with image pre-scaling to $256x256$ pixels. This forces the DF-Network to learn manipulation traces which are even included in downscaled images. 

Curiously enough, the metrics for DF-Net do sometimes even improve on the OSN-transmitted dataset versions. The effect is strongest for WhatsApp transmitted image data, where the average of the three metrics over all benchmark datasets increases by $0.00578$. This effect can also be found for Wu22 \cite{Wu2022}, where all metrics are higher for the Facebook transmitted Columbia dataset compared to the original (non-transmitted) data.

\section{Conclusion}
\label{sec:conclusion}
In this paper, we propose a lightweight network architecture for image manipulation detection. We share our model, the Digital Forensics Network (\href{https://zenodo.org/record/8142658}{DF-Net}, with the community. This model shows a better performance than several state-of-the-art methods on four well-established benchmark datasets. In particular, DF-Net outperforms its competitors for images transmitted over popular social networks such as Facebook or WhatsApp. With a simple and practical training concept, the DF-Net addresses the challenges of lossy operations (downscaling, filtering) and focuses on robust manipulation features. In extensive evaluations, we show that DF-Net has virtually no performance degradation on OSN-transmitted images, which is a unique feature compared to competitors in the field of image forgery detection. 
Furthermore, the detection speed is significantly higher than that of its closest competitor since DF-Net does not rely on a tiling process (see Tab.~\ref{tab:time-comparison}).


\section*{Acknowledgement}
\label{sec:acknowledgement}
\begin{tabular}{ll}
\raisebox{-.4\height}{\includegraphics[width=4cm]{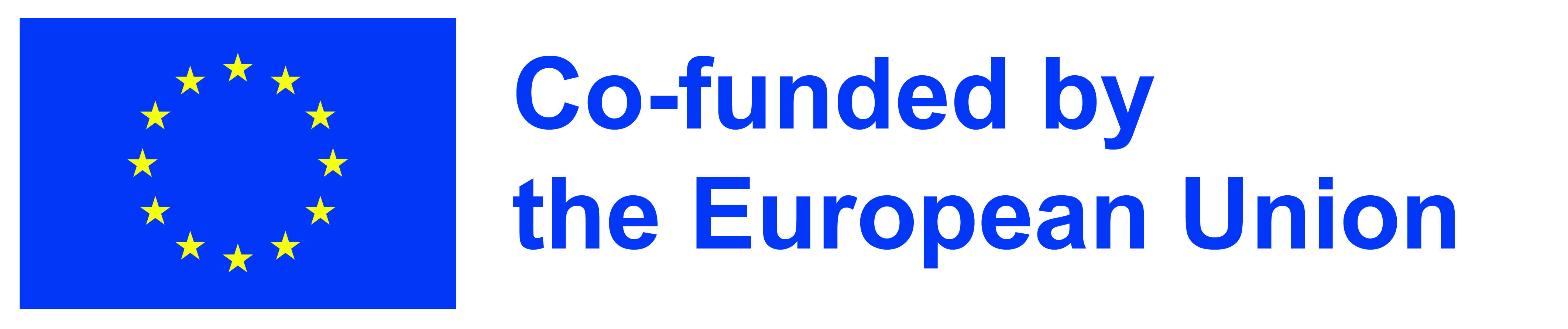}} & Project 101083573 — GADMO\\
\end{tabular}
\appendix

\bibliographystyle{apalike}


\end{document}